\documentclass[sigconf]{acmart}
\acmConference[CIKM '25]%
  {34th ACM International Conference on Information and Knowledge Management}%
  {November 10-14, 2025}%
  {Coex, Seoul, Korea}
\AtBeginDocument{%
  }

\copyrightyear{2025}
\acmYear{2025}
\setcopyright{acmlicensed}\acmConference[CIKM '25]{Proceedings of the 34th ACM International Conference on Information and Knowledge Management}{November 10--14, 2025}{Seoul, Republic of Korea}
\acmBooktitle{Proceedings of the 34th ACM International Conference on Information and Knowledge Management (CIKM '25), November 10--14, 2025, Seoul, Republic of Korea}
\acmDOI{10.1145/3746252.3761079}
\acmISBN{979-8-4007-2040-6/2025/11}

\newcommand{\myparagraph}[1]{\vspace{1mm} \noindent \textbf{#1}.}
\usepackage{float}
\usepackage{amsmath}
\usepackage{adjustbox}
\usepackage{multirow} 
\usepackage{makecell}
\usepackage{enumerate}
\usepackage{graphicx}
\usepackage{subcaption}
\begin{document}

\title{SGPT: Few-Shot Prompt Tuning for Signed Graphs}

\author{Zian Zhai}
\orcid{0009-0004-1430-3544}
\affiliation{%
  \institution{The University of New South Wales}
  \city{Sydney}
  \state{NSW}
  \country{Australia}
}
\email{zian.zhai@unsw.edu.au}

\author{Qing Sima}
\orcid{0009-0007-7595-123X}
\affiliation{%
  \institution{The University of New South Wales}
  \city{Sydney}
  \state{NSW}
  \country{Australia}
}
\email{q.sima@unsw.edu.au}

\author{Xiaoyang Wang}
\orcid{0000-0003-3554-3219}
\affiliation{%
  \institution{The University of New South Wales}
  \city{Sydney}
  \state{NSW}
  \country{Australia}
}
\email{xiaoyang.wang1@unsw.edu.au}

\author{Wenjie Zhang}
\orcid{0000-0001-6572-2600}
\affiliation{%
  \institution{The University of New South Wales}
  \city{Sydney}
  \state{NSW}
  \country{Australia}
}
\email{wenjie.zhang@unsw.edu.au}

\renewcommand{\shortauthors}{Zian Zhai et al.}

\begin{abstract}
Signed Graph Neural Networks (SGNNs) are effective in learning expressive representations for signed graphs but typically require substantial task-specific labels, limiting their applicability in label-scarce industrial scenarios.
In contrast, unsigned graph structures are abundant and can be readily leveraged to pre-train Graph Neural Networks (GNNs), offering a promising solution to reduce supervision requirements in downstream signed graph tasks. 
However, transferring knowledge from unsigned to signed graphs is non-trivial due to the fundamental discrepancies in graph types and task objectives between pre-training and downstream phases.
To address this challenge, we propose Signed Graph Prompt Tuning (SGPT), a novel graph prompting framework that adapts pre-trained unsigned GNNs to few-shot signed graph tasks. 
We first design a graph template based on balance theory to disentangle mixed node relationships introduced by negative links, mitigating the structural mismatches between unsigned and signed graphs.
We further introduce a task template that reformulates downstream signed tasks into a unified link prediction objective, aligning their optimization goals with the pre-training task.
Furthermore, we develop feature prompts that align downstream semantic spaces with the feature spaces learned during pre-training, and semantic prompts to integrate link sign semantics in a task-aware manner.
We conduct extensive experiments on seven benchmark signed graph datasets, demonstrating that SGPT significantly outperforms existing state-of-the-art methods, establishing a powerful and generalizable solution for few-shot signed graph learning.
\end{abstract}




\maketitle

\section{Introduction}
In many real-world social networks, interactions between entities involve conflicting relationships such as friendships versus enmities, trust versus distrust, or agreement versus disagreement~\cite{wu2020maximum}.
These opposing dynamics are naturally modeled as signed graphs, where each link encodes not only the existence of a connection but also its polarity: positive or negative.
Effectively understanding and modeling signed graphs is crucial for a wide range of applications, including predicting user behavior, recommender systems, community detection, and fraud detection~\cite{liu2021signed,DBLP:journals/tkde/SunCWZW22, li2024adarisk, wang2024efficient, wang2025effective}.
                                 
To learn expressive representations of signed graphs, Signed Graph Neural Networks (SGNNs) have attracted increasing research interest~\cite{derr2018signed,shu2021sgcl,huang2021sdgnn}. 
SGNNs incorporate link sign signals into the message-passing process, updating node embeddings by separately aggregating information from neighbors with different semantics. 
This approach has proven effective for various signed graph tasks, such as node classification~\cite{mercado2020signednc} and link sign prediction~\cite{huang2019sgat,huang2021sdgnn,li2023slgnn}.
Despite these advances, the application of SGNNs in real-world scenarios remains challenging due to limited labeled data and complex signed structures.
First, SGNNs are typically trained in a fully supervised manner, relying heavily on large quantities of labels, such as node classes and link signs.
However, in practice, collecting such annotations requires substantial human effort and is often infeasible at scale~\cite{wu2023surveyprompt,zhang2025label}.
As a result, SGNNs become vulnerable to overfitting and exhibit poor generalization performance in low-label scenarios ~\cite{tang2016signedsurvey,tang2016node,liu2021signed,zhang2025instance}.
Second, signed graph datasets tend to exhibit exceptionally sparse structures and higher levels of noise compared to unsigned graphs~\cite{zhang2024signedsurvey,shu2021sgcl,zhang2023sbgcl}. 
Learning robust SGNNs under such conditions is challenging, as models may overfit to spurious patterns rather than capturing meaningful semantics.

Inspired by the success of “pre-training and prompt tuning” in Large Language Models (LLMs)~\cite{brown2020language,shi2023don}, recent research has extended this paradigm to address label reliance and overfitting problems in conventional Graph Neural Networks (GNNs)~\cite{sun2022gppt,fang2024GPF}. 
In the pre-training phase, GNNs are trained on large-scale unsigned graphs using Self-Supervised Learning (SSL) objectives to learn task-agnostic structural priors~\cite{hu2019linkpred,you2021graph}.
In the downstream phase, lightweight prompts are introduced as task-specific adapters and optimized using limited labels, while keeping the pre-trained GNN backbone frozen.
This approach has demonstrated strong performance with minimal supervision and excellent parameter efficiency~\cite{sun2023all,liu2023graphprompt}, offering a promising solution to alleviate label dependency and overfitting in supervised SGNNs.

However, existing prompt-tuning methods are designed exclusively for unsigned graphs and overlook downstream scenarios where tasks involve signed relational information.
Thus, in this paper, we address this gap and focus on transferring pre-trained unsigned GNNs to signed graph tasks via prompt tuning. 
This problem is non-trivial due to two key challenges that are absent in prior graph prompt-tuning works.

The first challenge lies in \textit{the divergence between graph types used in pre-training and downstream tasks.}
In pre-training, unsigned graphs are utilized where links uniformly represent connectivity~\cite{hu2019linkpred}, and GNNs are trained under the homophily assumption, aggregating features from all neighboring nodes without distinguishing their relational semantics~\cite{mcpherson2001birds}.
In contrast, downstream signed graphs contain both positive and negative links that encode inherently conflicting relationships.
This semantic polarity fundamentally violates the homophily assumption of the message passing mechanism in pre-trained GNNs, as uniformly aggregating features from semantically opposite neighbors can result in misleading node embeddings. 
Existing prompt-tuning methods, such as GraphPrompt+\cite{yu2024graphprompt++}, GPPT~\cite{sun2022gppt}, GPF~\cite{fang2024GPF}, and ProG~\cite{sun2023all}, assume consistent link semantics in both pre-training and downstream graphs and therefore lack the capacity to model link polarities in signed graphs.
Therefore, dedicated efforts are needed to narrow down the graph-type divergence.

Another challenge is \textit{the divergence between task objectives in pre-training and downstream tasks.}
Pre-training on unsigned graphs typically adopts SSL objectives that focus on preserving instinct graph structures such as graph similarities and node connectivity~\cite{zhang2018link,you2021graph}.
In contrast, downstream signed tasks aim to optimize the task-specific loss function, e.g., fit the ground truth of node labels or link sign labels~\cite{derr2018signed,huang2019sgat}, where models must capture not only graph structure but also handle the complexity introduced by signed relational semantics. 
This misalignment between general pre-training objectives and semantics-related downstream tasks poses a significant barrier to effectively applying pre-trained GNNs, especially under low-label conditions. 
Although recent prompt-tuning methods, e.g., HGPrompt~\cite{yu2024hgprompt}, ProG, and GraphPrompt+ address divergences between pre-training tasks and few-shot graph tasks, they fail to account for link polarity, making them inadequate for tasks like link sign prediction that demand explicit modeling of relational semantics.

To address these challenges, we propose \textbf{\underline{S}}igned \textbf{\underline{G}}raph \textbf{\underline{P}}rompt \textbf{\underline{T}}uning (\textbf{SGPT}), a novel transfer learning framework to adapt the pre-trained unsigned GNNs to signed graph tasks under few-shot settings.
SGPT bridges the gaps between pre-training and downstream phases through two unification templates, i.e., graph and task templates. 
Specifically, to mitigate the divergence in graph types, we propose a novel graph template based on the balance theory to address the polarity of node relationships in signed graphs.
This template disentangles multi-hop node relationships into three distinct graph channels, i.e., positive, negative, and topological, each capturing a unique type of relational semantics while maintaining compatibility with the homophily assumption in pre-trained unsigned GNNs. 
To reduce the divergence in task objectives, we introduce a task template that reformulates the signed tasks into a unified link prediction format, thereby ensuring consistent optimization objectives across both phases. 
Furthermore, we develop two lightweight and semantic-aware prompts, i.e., feature prompts and semantic prompts. 
Feature prompts adaptively modify the downstream semantic space of each graph channel, aligning representations from different semantics with the pre-trained feature distribution. 
Semantic prompts integrate multi-channel outputs into a final representation by weighting each channel’s contribution based on the nature of specific downstream tasks.

To sum up, the main contributions of this paper are as follows:
\begin{itemize}
    \item To the best of our knowledge, this is the first study to explore prompt learning for transferring knowledge from pre-trained unsigned GNNs to few-shot signed tasks, a development with wide applicability to real‐world social network analysis.
    \item We propose a balance theory-based graph template that decomposes signed link polarity into semantically separated channels, effectively bridging the divergence between unsigned and signed graphs.
    \item We develop feature and semantic prompts tailored for signed graphs, which align input feature spaces and integrate polarity signals in a task-aware manner.
    \item We conduct comprehensive experiments on seven benchmark signed graphs, demonstrating that SGPT significantly outperforms existing supervised and prompt-tuning methods in low-label settings.
\end{itemize}

\section{Related Work}
\myparagraph{Signed graph neural networks}
Recently, with the development of GNNs, researchers begin applying these powerful techniques to learn low-dimensional representations for signed graphs.
For instance, SGCN~\cite{derr2018signed} adopts balance theory and designs a path-based feature aggregation and propagation mechanism for undirected signed graphs.
SLGNN~\cite{li2023slgnn} designs graph convolution filters to extract low- and high-frequency signals on positive and negative links, and combines them into the unified message-passing of GNNs.
SiGAT~\cite{huang2019sigat} introduces GAT to directed signed graphs and designs a motif-based GNN with social theory to model the node relationships.
SDGNN~\cite{huang2021sdgnn} further uses sociological theories to reconstruct link signs, directions, and signed directed triangles for expressive embeddings.
Although these SGNNs effectively capture the semantics of signed relationships, they are trained in a fully supervised manner and rely heavily on task-specific labels for supervision.

\myparagraph{Graph pre-training}
The recent success of LLMs inspires researchers to apply the pre-training strategy on GNNs to enhance their generalization capabilities~\cite{tan2025hydra}. 
Essentially, graph pre-training leverages the easily accessible graph structure data to train a GNN in a self-supervised manner, e.g., link prediction~\cite{hu2019linkpred}, node attribute reconstruction~\cite{grover2016node2vec}, and graph pattern discrimination~\cite{you2021graph}. 
The pre-trained GNNs learn universal, transferable patterns and serve as an effective initialization for different downstream tasks.
They are subsequently fine-tuned with the specific downstream objectives. 
For signed graphs, although there are some studies that utilize contrastive learning to train SGNNs, e.g., SGCL~\cite{shu2021sgcl} and SBGCL~\cite{zhang2023sbgcl}, they still rely on the link sign labels for graph augmentation and supervision.
As a result, these models are unsuitable for pre-training SGNNs, particularly for few-shot link sign prediction scenarios.

\myparagraph{Graph prompt tuning}
In LLMs, prompt tuning bridges the gap between pre-training and downstream tasks by introducing tunable lightweight tokens and reformulating downstream tasks to align with the pre-training objectives~\cite{sun2023all, tan2025paths}. 
This approach maximizes the utilization of pre-training knowledge, thereby reducing the supervision requirements in the downstream phase.
Recent research has adopted this strategy to relieve the label reliance in supervised GNNs.
For example, GPPT~\cite{sun2022gppt} reformulates node classification tasks to link prediction, but its design is limited to that specific task.
GraphPrompt~\cite{liu2023graphprompt} introduces a subgraph similarity objective as the pre-training task and extends prompt designs to graph classification tasks.
Building on this, GraphPrompt+~\cite{yu2024graphprompt++} inserts prompt vectors within every layer of the pre-trained GNN encoder to better leverage hierarchical information.
ProG~\cite{sun2023all} reformulates different downstream problems to the graph-level tasks and learns prompt tokens via meta-learning.
HGPrompt~\cite{yu2024hgprompt} transfers the pre-trained GNNs to heterogeneous tasks by unifying not only the pre-training and downstream tasks but also the homogeneous and heterogeneous graphs.
While these methods demonstrate the effectiveness of prompt tuning in graph domains, they are exclusively designed for unsigned or heterogeneous graphs and fail to address signed relational semantics, which is critical for signed graph tasks. 

\section{Preliminary}
%For the convenience of presentation, we first introduce the definitions and notations used in this paper.

\myparagraph{Signed graph} A signed graph is denoted as ${\mathcal G} = (\mathcal{V}, {\mathcal{E}^{+}}, \mathcal{E}^{-})$, where $\mathcal{V} = \{v_1, v_2, ..., v_n\}$ denotes the set of $n$ nodes. 
${\mathcal{E}^{+}}$ and $\mathcal{E}^{-}$ represent the sets of positive and negative links respectively, 
such that ${\mathcal{E}^{+}}\cap \mathcal{E}^{-} = \varnothing$. 
We use $A^{+}\in \mathbb{R}^{n\times n}$ and $A^{-}\in \mathbb{R}^{n\times n}$ to represent the positive and negative binary adjacency matrices. 
$X \in \mathbb{R}^{n\times d_{\text{in}}}$ is the feature matrix, where $d_{\text{in}}$ is the dimension of input features. 
Similarly, an {unsigned graph} disregards link signs, which can be represented as ${\mathcal G} = (\mathcal{V}, {\mathcal{E}})$. An adjacency matrix $A$ can be constructed where $a_{ij} = 1$ if $v_i$ connects to $v_j$, otherwise $a_{ij} = 0$.

\myparagraph{Balance theory}
Balance theory is a fundamental concept in social psychology, focusing on the stability of triadic structures in human relationships.
The core concept is that a triad is considered \textit{balanced} if it contains an even number of negative
links; otherwise, it is \textit{unbalanced}. 
%The two triangles on the left in Fig.~\ref{fig:triangle} are balanced triangles as they contain even number of negative links. The two triangles on the right are unbalanced because the number of negative links is odd.
This principle can be abstracted as a well-known proverb: “the friend of my friend is my friend, and the enemy of my enemy is my friend”~\cite{cartwright1956structural}.
In this paper, we develop a polarity disentanglement strategy based on balanced structures, which are more prevalent than unbalanced ones in real-world social networks~\cite{shu2021sgcl}.

\myparagraph{Graph Neural Networks (GNNs)}
GNNs learn the node representation by recursively aggregating features from neighbors, also known as \textit{message-passing}~\cite{sun2022gppt, tan2023higher, li2024efficient}. Formally, the representation of node $v$ at the $l$-th layer is:
\begin{equation}
    \text{h}^{(l)}_v = \text{AGG}(\{\text{h}_u^{(l-1)}, u \in \mathcal{N}(v) \cup v\}, \theta^{(l)}),
\end{equation}
where $\text{h}^{(0)}_v$ is the initial node feature, $\mathcal{N}(v)$ is the neighbor set of node $v$, and $\theta^{(l)}$ is the parameters of the $l$-th layer of the GNN. 
The aggregation function $\text{AGG}(\cdot)$ combines the embedding of node $v$ and its neighbors, which is typically sum, mean, or max pooling. 

\myparagraph{Problem statement}
In this paper, we focus on adapting the pre-trained unsigned GNNs to two widely studied signed graph tasks, namely node classification (\textbf{NC}) and link sign prediction (\textbf{LSP}). 
In particular, we consider few-shot settings, due to the scarcity of labels in real-world scenarios. 
\begin{itemize}
    \item Few-shot node classification. Let $C = \{c_1, c_2, ..., c_m\}$ denote the set of node classes. 
    Each node in the signed graph $\mathcal{G}$ is associated with a label $c\in C$. In the few-shot setting, the training set includes only a small number of nodes per class, significantly fewer than the total number of nodes.
    Our goal is to predict the unknown node labels in the test set given that all link signs are known.
    
    \item Few-shot link sign prediction. Let $C = \{P, N\}$ denote the positive and negative link signs. 
    Each linked node pair in the signed graph $\mathcal{G}$ has a link sign $c \in C$. We mask a portion of the link signs in $\mathcal{G}$ and only have a small number of link instances for both positive and negative classes in the training set, which is significantly fewer than the total number of linked node pairs.
    In the test set, we already know the queried node pairs are connected, and our task is to predict their signs.
\end{itemize}

\section{Proposed Approach}

In this section, we present SGPT, a novel prompt tuning framework designed for few-shot signed tasks.
We first introduce the overall framework and graph pre-training stage (Sections 4.1 and 4.2).
Next, we describe the graph and task templates, which unify signed and unsigned graphs and bridge pre-training tasks and downstream signed tasks (Section 4.3).
Finally, we introduce feature prompts and semantic prompts, which align the feature spaces of distinct semantic channels with the pre-training task and integrating semantic information in a task-aware manner (Section 4.4).

\subsection{Overall Framework}
\begin{figure*}[t]
  \centering
\includegraphics[width=\linewidth]{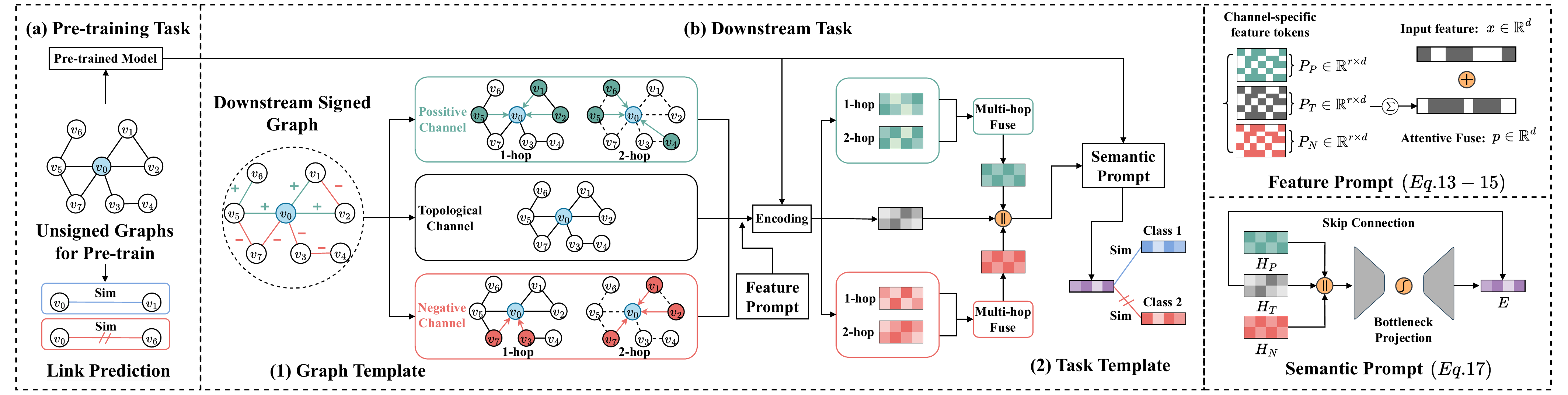}
  
%\vspace{-5mm}
  
  \caption{Overall framework of SGPT.}
%\vspace{-5mm}
  \label{fig:frame}
\end{figure*}
In this subsection, we present the overall framework of SGPT, as shown in Fig. \ref{fig:frame}. 
At the pre-training stage in Fig.~\ref{fig:frame}(a), we use link prediction (\textbf{LP}) task on unsigned graphs to pre-train an unsigned GNN. 
During the downstream stage in Fig.~\ref{fig:frame}(b), we design the graph and  task templates specifically for signed graphs to unify the pre-training and downstream tasks.
First, the graph template disentangles the mixed node relationships in signed graphs. 
It leverages the balance theory to extract multi-hop node relationships and generate graph samples across three channels, i.e., positive, negative, and topological, each with a consistent type of internal link semantics and satisfying the homophily assumption of the pre-trained unsigned GNN. 
These segregated samples are then fed into the pre-trained GNN in parallel to generate their corresponding representations.
Second, a task template unifies the pre-training and downstream tasks.
It reformulates link sign prediction and node classification into the same form as link prediction, ensuring a consistent optimization objective.
With unified graph types and task forms, we further design the feature and semantic prompts as tunable lightweight parameters to adapt the pre-trained GNN.
The feature prompts are added to the input node features of segregated graph samples before the encoding process to align the downstream semantic space with that of the pre-training task. 
Afterward, the semantic prompt adaptively integrates the embeddings of graph samples, generating the representations tailored to the requirements of downstream tasks.

\subsection{Graph Pre-Training}
In this paper, we adopt link prediction as the pre-training task~\cite{hu2019linkpred}. 
Let $\mathcal{G}^{\text{pre}}$ denote the pre-training graph with node feature matrix $X \in \mathbb{R}^{n\times d_{\text{in}}}$ and the adjacency matrix $A^{\text{pre}}$, where a portion of links are masked as the training set, and the corresponding elements in the adjacency matrix are set to $a_{ij} = 0$. 
The node embedding matrix can be generated by the GNN based on the unmasked links, denoted as $H = f_{\theta}^{\text{pre}}(\mathcal{G}^{\text{pre}}, X)$, where $H \in \mathbb{R}^{n\times d_{\text{out}}}$, and $d_{\text{out}}$ is the dimension of the output embeddings.

Given a node $v$ in the pre-training graph $\mathcal{G}^{\text{pre}}$, we randomly sample a positive node $a$ from $v$'s neighbors, and a negative node $b$ from the graph which does not link to $v$, forming the triplet $(v, a, b)$ in the training set $\mathcal{D}_\text{pre}$.
The objective of LP is to predict higher similarity to $v$ and $a$ while giving lower similarity to $v$ and $b$. 
The GNN is optimized to minimize the loss function:
% $\mathcal{L}_{\text{pre}} = $
\begin{align}
    \mathcal{L}_{\text{pre}} =
    -\sum_{(v, a, b) \in \mathcal{D}_{\text{pre}}} \ln \frac{\exp\left(\text{Sim}(h_v, h_a)\right)}{\sum_{u \in \{a, b\}} \exp\left(\text{Sim}(h_v, h_u)\right)} \label{eq:preloss},
\end{align}
where $\text{Sim}(h_v, h_u)$ denotes the similarity of node embeddings, which can be calculated using various similarity measures, such as cosine similarity.

\subsection{Template Design}

\myparagraph{Graph template} 
Negative links in signed graphs introduce more complex node relationships and violate the homophily assumption in pre-trained GNNs. 
To address the divergence in graph types, we start from extracting link semantics and propose a graph template.
Based on the balance theory~\cite{heider1946attitudes}, it explicitly extracts node relationships at different hops to generate multiple graph samples. Each maintains a consistent type of internal link semantics, i.e., positive, negative and topological, segregating the mixed relationships in the original signed graphs.

According to the balance theory, a path is balanced if the number of negative links in it is even, otherwise the path is unbalanced. 
% while an unbalanced path has an odd number of negative links. 
%The balanced paths indicate positive relationships between the source node and the target node while unbalanced paths represent negative relationships~\cite{derr2018signed}. 
If a source node and a target node are connected by a balanced (resp. unbalanced) path, they may show explicit or implicit positive (resp. negative) relation~\cite{derr2018signed,DBLP:journals/tkde/SunWWCZL24}.
%Given this principle, the graph template explicitly extracts node relationships.
%For nodes with balanced (unbalanced) paths, it manually connects them based on their hop count to generate multiple positive (negative) graph samples. 
Formally, given the downstream signed graph ${\mathcal G}$ with positive and negative adjacency matrix $A^{+}$ and $A^{-}$, the $1$-hop positive and negative graph samples are generated by directly selecting positive and negative neighbors of each node. 
The adjacency matrices of $1$-hop graph samples can be denoted as:
\begin{equation}
    A^1_P = A^+,    A_N^1 = A^-.
\end{equation} 

\begin{figure}
\centering
  \includegraphics[width=0.4\textwidth]{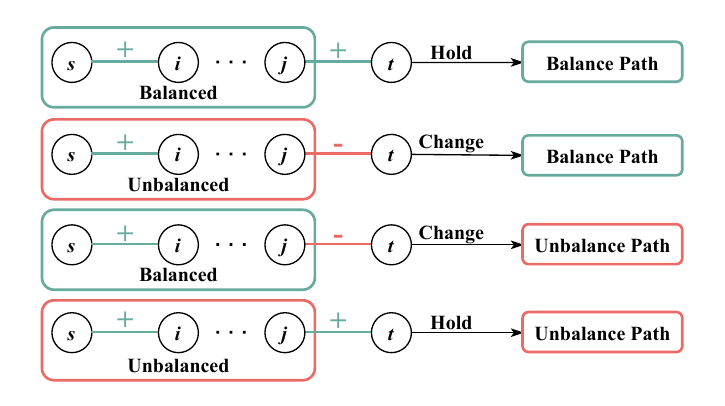}
%\vspace{-3mm}
  \caption{Balanced and unbalanced paths in signed graphs.}
%\vspace{-5mm}
  \label{fig:path}
\end{figure}

Furthermore, we consider multi-hop neighbors to capture the global information. 
For $k$-hop ($k\geq 2$) graph samples, adding a negative link to a balanced path makes it unbalanced, while an unbalanced path with an additional negative link becomes balanced. 
Conversely, adding a positive link preserves the path's balance status, as shown in Fig.~\ref{fig:path}.
% a balanced path with a negative link results in an unbalanced path, while an unbalanced path with a negative link becomes balanced. 
% Conversely, adding a positive link to a balanced path leads to another balanced path, whereas adding a positive link to an unbalanced path results in an unbalanced path. 
Thus, the adjacency matrices of $k$-hop positive and negative graph samples are calculated by:
\begin{equation}
A^k_P = I(A_P^{k-1}A^1_P + A_N^{k-1}A^1_N),
\end{equation}
\begin{equation}
A^k_N = I(A_P^{k-1}A^1_N + A_N^{k-1}A^1_P),
\end{equation}
where $I(\cdot)$ is an indicator function by setting non-zero elements to 1. 
In this way, we explicitly extract the relationships between each node and its 1- to $k-$ hop neighbors, thereby generating $k$ samples for both the positive and negative semantics. 
Each sample contains links that represent a consistent type of node relationship, either positive or negative.

For simplicity, we refer to the set of positive samples as the positive channel, and the set of negative samples as the negative channel, which is denoted as: 
\begin{align}
    \mathcal{G}_P &= \{\mathcal{G}^1_{P}, \ldots, \mathcal{G}^k_{P} : \mathcal{G}^i_{P}=(X, A^i_P) \text{ and } 1\leq i \leq k\}, \\
    \mathcal{G}_N &= \{\mathcal{G}^1_{N}, \ldots, \mathcal{G}^k_{N} : \mathcal{G}^i_{N}=(X, A^i_N) \text{ and } 1\leq i \leq k\}.
\end{align} 

In addition, to preserve the original graph topology, we mask the link signs in the downstream signed graph and convert it to an unsigned sample in the topological channel. 
In this channel, links represent only the connection between nodes, which is denoted as: 
\begin{equation}
    \mathcal{G}_T = (X, A^{+}\cup A^{-}).
\end{equation}

To sum up, the graph template segregates mixed link semantics in the downstream signed graph into multiple samples across three channels. 
Each channel represents a consistent type of link semantics, denoted as $\mathcal{S} = \{P, N, T\}$, and this process is represented as $\mathcal{GT}(\mathcal{G}) =  \{\mathcal{G}_P, \mathcal{G}_N, \mathcal{G}_T\}$.
%We name this process as \textit{graph type prompt}, denoted as:
%\begin{equation}
    %\mathcal{GTP}(\mathcal{G}) = \{\mathcal{G}^o\cup \mathcal{G}_{P}\cup \mathcal{G}_{N}\}.
%\end{equation}
%

\myparagraph{Task template}
The task template converts the downstream link sign prediction and node classification tasks to the same form used in the pre-training link prediction.
%We first introduce the template of link prediction, followed by the templates of link sign prediction and node classification.
\begin{itemize}
\item \underline{Link prediction}: 
Given a pre-training unsigned graph and a node triplet $(v, a, b)$, there exists a link between $v$ and $a$, whereas $v$ and $b$ are not connected.
With the optimization objective in Equation \ref{eq:preloss}, the model is expected to assign higher similarity to linked node pairs than unlinked ones, which is denoted as:
\begin{equation}
    \text{Sim}(h_v, h_a) > \text{Sim}(h_v, h_b).
\end{equation}

\item \underline{Link sign prediction}: Given a downstream signed graph with link sign classes $C = \{P, N\}$, we introduce one prototype into the link embedding space for each link label class. 
Their embeddings are defined as $E = [e_{P}, e_{N}] \in \mathbb{R}^{2 \times 2d_{\text{out}}}$. 
Here, we use the concatenation of the embeddings of node $v$ and node $u$ in the queried link as the link embedding, which is denoted as $e_{(v, u)} = [h_v \| h_u]$.
Then, the model predicts the link sign by selecting the prototype with the highest similarity to the queried link:
\begin{equation}
c_{(v, u)} = \arg\max_{c \in \{P, N\}} \text{Sim}(e_{(v, u)}, e_c).
\end{equation}

\item \underline{Node classification}: Given a downstream signed graph with $m$ node classes $C = \{c_1, ..., c_m\}$, we adopt $m$ class prototypes. 
Their embeddings are defined as $E = [e_{c_1}, ..., e_{c_m}] \in \mathbb{R}^{m\times d_{\text{out}}}$. 
A node can then be classified by selecting the class with the highest similarity:
\begin{equation}
    c_{v} = \arg\max_{c_i \in C} \text{Sim}(h_{v}, e_{c_i}).
\end{equation}
\end{itemize}

\subsection{Prompt Design}

\myparagraph{Feature prompt}
%In our target scenario, we aim to leverage the model pre-trained with LP for few-shot LSP and NC. 
%Different downstream tasks rely on different task-specific features, which may not align well with the task-agnostic features learned during pre-training~\cite{yu2024hgprompt}. 
Due to the different natures of pre-training and downstream tasks, they naturally prioritize different aspects of input features~\cite{sun2023all,ma2024hetgpt}. To effectively utilize the pre-trained model's knowledge, it is imperative to align downstream input features with the pre-training task.
%Moreover, retraining the model with limited task-specific labels can easily overfit and cause catastrophic forgetting \cite{sun2019meta,yap2021addressing}.
To this end, inspired by the success of visual prompts in the computer vision domain \cite{wu2022unleashing}, we introduce the feature prompt~\cite{fang2024GPF}.
It involves adding lightweight vectors to modify the input node features before the GNN encoding, acting as task-specific feature augmentation to align the downstream input space with that of the pre-training task.
In the downstream stage, only these prompts are updated while the parameters of the pre-trained model stay frozen.

Prior research~\cite{sun2023all, fang2024GPF} has demonstrated that there always exists a feature prompt vector, which is equivalent to a graph transformation function $t_{\xi}$, satisfying the following:
\begin{equation}
f_{\theta}^{\text{pre}}(\mathcal{G}^{\text{pre}}, X + p^*) = f_{\theta}^{\text{pre}}(t_{\xi}(\mathcal{G}^{\text{pre}}, X)) + O_{p\theta},
\end{equation}
where $O_{p\theta}$ represents the error bound and can be reduced by tuning the feature prompt $p$.
This perspective indicates the feasibility and efficiency of modifying the input feature space to learn the graph transformation function, thereby enabling the dynamic adjustment of node representations to meet the specific requirements of different downstream tasks.

However, since signed graph tasks tend to emphasize distinct features for different types of link semantics~\cite{shu2021sgcl,huang2021sdgnn}, we design feature prompts tailored separately for each semantic channel.
Formally, for the downstream signed graph, we assume that we obtain graph samples of three channels via the graph template $\mathcal{GT(G)}$. 
For each channel with a type of link semantics $s \in \mathcal{S}$, we employ a feature prompt $P_s$ consisting of $r$ independent basis vectors. 
Each basis vector has the same dimension of $d_{\text{in}}$ as the input node features. 
The feature prompt $\mathcal{P}$ is denoted as:
\begin{equation}
    \mathcal{P} = \{P_s | s \in \mathcal{S}\},  
    \mathcal{P}_s = \{p_s^1, ..., p_s^r\}.
\end{equation}

For each node $v$ in the graph sample $\mathcal{G}^i_s$, we add the attentive aggregation of the prompt basis in $P_s$ to its input feature $x$. Then the prompted feature $x^i_s$ can be obtained by:
\begin{equation}
    x^i_{s} = x + \sum_{j=1}^r \alpha _{s}^j \cdot p_{s}^j,
\end{equation}
\begin{equation}
    \alpha _{s}^j = \frac{\text{exp}((q_{s}^j)^\top
 x)}{\sum ^r_{t=1}\text{exp}((q_{s}^t)^\top x)},
\end{equation}
where $q_{s}^j$ is the linear projection head for the prompt basis $p_{s}^j$ to calculate the attention weight. 
Subsequently, we pass prompted node features of each graph sample $\mathcal{G}_s^i$, represented as $X^i_s$, to the pre-trained unsigned GNN model to obtain the prompted node embedding matrix: $H_s^i = f^\text{pre}_\theta(\mathcal{G}_s^i, X^i_s)$.

As downstream tasks may emphasize information from different levels, i.e., the local structure contained from closer neighbors or global information from further nodes, we utilize a learnable hop-wise coefficient $w_s \in \mathbb{R}^{1\times K}$ for both positive and negative channels to fuse the prompted embedding of graph samples with different hops. 
The fused embeddings of positive and negative channels are represented as:
\begin{equation}
    H_P = \sum _{i=1}^kw_P^iH_P^i, 
    H_N = \sum _{i=1}^kw_N^iH_N^i.
\end{equation}

Since we do not extract multi-hop relationships for the graph sample in the topological channel, we pass its prompted node features $X_T$ to the GNN and only use the prompted embedding $H_T =  f^\text{pre}_\theta(\mathcal{G}_T, X_T)$ for the topological channel.

\myparagraph{Semantic prompt} 
After obtaining the prompted embeddings of three channels, we design a semantic prompt to adaptively aggregate them based on the specific requirements of downstream tasks. 
A naive approach is to concatenate the embeddings. 
However, such aggregation assumes equal contribution from all channels and may fail to capture their varying importance.
To address this limitation, we adopt a lightweight adapter module as the semantic prompt, which preserves both the original node embeddings and the semantic distinctions in a task-aware manner.

Formally, an adapter includes a down projection $
\textbf{W}_{\text{down}} : \mathbb{R}^{3d_{\text{out}}} \to \mathbb{R}^{d_{\text{mid}}}
$, a ReLU activation function, and an up projection $\textbf{W}_{\text{up}} : \mathbb{R}^{d_{\text{mid}}} \to \mathbb{R}^{d_{\text{out}}}$. 
To reduce the learnable parameters, the mid-dimension $d_{\text{mid}}$ is usually set to a small number as a bottleneck, i.e., $d_\text{mid} \ll d_\text{out}$. 
With the adapter, the integrated node embedding matrix is calculated as follows:
\begin{equation}
    E = H_T + \text{BN}(\textbf{W}_{ \text{up}}(\text{ReLU}(\textbf{W}_{\text{down}}(H_T|| H_P || H_N)))),
\end{equation}
where BN denotes batch normalization.
The addition between $H_T$ and the adapter output serves two purposes.
First, $H_T$ preserves the topology without the influence of link signs, providing a neutral representation that captures the graph structure~\cite{perozzi2014deepwalk}. 
It is then enriched by the adapter's integration of the semantic channels.
Second, this addition acts as a part of a residual network, preventing gradient vanishing~\cite{houlsby2019adapter,li2024adaptergnn}.

More complex prompts, such as a linear transformation matrix or an attention layer, can be alternatives.
In few-shot scenarios, prompt designs with fewer parameters are preferred due to their lower tuning difficulties and better generalization abilities~\cite{yu2024generalized}. Therefore, we adopt the adapter here for its lightweight nature and tuning efficiency.

\myparagraph{Prompt tuning}
Given the training set $\mathcal{D}_{\text{down}} = \{(x_1, y_1),$ $ (x_2, y_2),...\}$, where each $x_i$ can be either a node or a linked node pair instance, and $y_i$ is the corresponding label $y_i \in Y$.
With the parameters of pre-trained GNNs frozen, the prompts are optimized by minimizing the loss function:
% $\mathcal{L}_{\text{down}} = $
\begin{align} 
    \mathcal{L}_{\text{down}} =
     -\sum_{(x_i,y_i) \in \mathcal{D}_{\text{down}}} \ln \frac{\exp\left(\text{Sim}(e_{x_i}, e_{y_i})\right)}{\sum_{c \in Y} \exp\left(\text{Sim}(e_{x_i}, e_{y_c})\right)},
\end{align}
where $e_{x_i}$ can be either the node or link embedding, and $e_{y_c}$ is the embedding of the class prototype. 

\section{Experiments}
In this section, we conduct extensive experiments on seven datasets to assess the performance of SGPT.
We first present the experimental setup.
Afterward, we evaluate the models' performance on few-shot NC and LSP tasks.
Furthermore, we conduct ablation study to evaluate the impacts of each proposed component.
Then, we conduct efficiency and complexity analysis.
Finally, we investigate the sensitivity of hyper-parameters in SGPT.
\subsection{Experimental Setup}

\myparagraph{Datasets} 
We conduct experiments on four datasets for link sign prediction (i.e., \textit{Bitcoin-Alpha}, \textit{Bitcoin-OTC}~\cite{kumar2016edge}, \textit{Slashdot}, \textit{Epinions}~\cite{leskovec2010signed}) and three node classification datasets (i.e., \textit{Wikipedia-RfA}, \textit{Wikipedia-Elec}~\cite{mercado2020node}, Wikipedia-Editor~\cite{kumar2015vews}) to evaluate our framework.
The statistics of datasets are presented in Table \ref{tab:statistics}. 
\begin{itemize}
        \item \textit{Bitcoin-Alpha} and \textit{Bitcoin-OTC} are who-trust-whom networks of people who use Bitcoin on the platforms of Bitcoin Alpha and Bitcoin OTC. Since Bitcoin users are anonymous, each user marks trust or distrust tags to others to ensure the security of the trading.
        \item \textit{Epinions} is also a who-trust-whom online social network of a general consumer review site. Members of the site can decide whether to trust each other's reviews.
        \item \textit{Slashdot} is a technology-related news website known for its specific user community, which allows users to mark each other as friends or foes.
        \item \textit{Wikipedia-RfA} and \textit{wikipedia-Elec} are social networks of editors on Wikipedia. The editors request to be administrators, and any other Wikipedia members may give a supporting, neutral, or opposing vote. The labels of nodes represent whether the editors are elected as administrators.
        %\item \textit{Wikipedia-Editor} is extracted from UMD Wikipedia where there are vandal and benign editors of Wikipedia. Each node is labeled as vandal or benign. Positive and negative links represent whether the node pairs belong to the same category.
        \item \textit{Wikipedia-Editor} is from UMD Wikipedia where there are vandal and benign editors of Wikipedia. Each node is labeled as vandal or benign. Positive and negative links represent whether the node pairs belong to the same category.
    \end{itemize}
\begin{table}[t]
\centering
\fontsize{9pt}{10pt}\selectfont
\setlength{\tabcolsep}{1mm}
\caption{Dataset statistics.}
\begin{tabular}{@{}l|l|cccccc@{}}
\toprule
Task & Dataset & \makecell{\# Nodes} & \makecell{\# Pos \\ Links} & \makecell{\# Neg \\ Links} & \makecell{\# Node \\ Classes}\\ \midrule
\multirow{4}{*}{LSP} & Bitcoin-Alpha & 3,783 & 22,650 & 1,536 & - \\
                                      & Bitcoin-OTC & 5,881 & 32,029 & 3,563 & - \\
                                      & Slashdot & 81,871 & 422,349 & 123,322 & - \\
                                      & Epinions & 131,828 & 717,667 & 123,705 & - \\ 
                                      \midrule
\multirow{3}{*}{NC} & WikiElec & 7,194 & 158,575 & 44,807 & 2 \\
                                     & WikiRfA & 11,393 & 267,761 & 78,456 & 2 \\ 
                                     & WikiEditor & 21,535 & 538,502 & 158,008 & 2 \\
                                     \midrule           
                                \bottomrule
\end{tabular}
%\vspace{-2mm}
%\vspace{-5mm}
\label{tab:statistics}
\end{table}

\begin{table*}[!t]
\centering
\setlength{\tabcolsep}{1mm}
\fontsize{9pt}{10pt}\selectfont
\renewcommand{\arraystretch}{1.0} % 调整表格行间距
\caption{Performance on LSP and NC in ROC-AUC score (\%, mean$\pm$std). The best performance is in bold.}
\begin{tabular}{cc|cccc|ccc}
\toprule
\multicolumn{2}{c|}{\multirow{2}{*}{\textbf{Methods}}} & \multicolumn{4}{c|}{\textbf{Link Sign Prediction}} & \multicolumn{3}{c}{\textbf{Node Classification}} \\ 
\multicolumn{2}{c|}{} & Bitcoin-Alpha & Bitcoin-Otc & Epinions & Slashdot & Wikipedia-Elec & Wikipedia-RfA & Wikiedia-Editor \\ 
\midrule
\midrule
\multirow{2}{*}{GNN} & GCN & 71.28\scriptsize{$\pm$1.39} & 70.77\scriptsize{$\pm$2.34} & 74.12\scriptsize{$\pm$0.73} & 55.48\scriptsize{$\pm$1.54} & 68.70\scriptsize{$\pm$10.59} & 62.77\scriptsize{$\pm$17.74} & 52.50\scriptsize{$\pm$2.24} \\ 
 & GAT                      & 67.25\scriptsize{$\pm$2.44} & 73.04\scriptsize{$\pm$1.30} & 68.78\scriptsize{$\pm$2.99} & 56.16\scriptsize{$\pm$1.83} & 51.11\scriptsize{$\pm$\phantom{0}8.56} & 56.64\scriptsize{$\pm$\phantom{0}3.78} & 48.92\scriptsize{$\pm$4.14} \\
\midrule
\multirow{2}{*}{SGNN} & SGCN & 66.87\scriptsize{$\pm$1.95} & 65.17\scriptsize{$\pm$1.78} & 68.51\scriptsize{$\pm$1.75} & 57.79\scriptsize{$\pm$1.06} & 51.06\scriptsize{$\pm$\phantom{0}1.52} & 51.03\scriptsize{$\pm$\phantom{0}0.68} & 48.51\scriptsize{$\pm$0.66} \\ 
& SDGNN                      & 74.48\scriptsize{$\pm$1.54} & 72.69\scriptsize{$\pm$2.52} & 71.62\scriptsize{$\pm$0.90} & 54.15\scriptsize{$\pm$1.08} & 56.85\scriptsize{$\pm$18.52} & 54.15\scriptsize{$\pm$\phantom{0}4.19} & 50.39\scriptsize{$\pm$ 1.61} \\
\midrule
\multirow{2}{*}{FT} & LinkPred                      & 73.29\scriptsize{$\pm$2.17} & 73.41\scriptsize{$\pm$1.70} & 78.47\scriptsize{$\pm$0.40} & 57.25\scriptsize{$\pm$1.34} & 49.68\scriptsize{$\pm$23.85} & 51.53\scriptsize{$\pm$22.62} & 49.03\scriptsize{$\pm$6.22} \\ 
& GraphCL                      & 69.59\scriptsize{$\pm$3.43} & 71.57\scriptsize{$\pm$4.01} & 69.43\scriptsize{$\pm$5.46} & 60.04\scriptsize{$\pm$1.79} & 63.66\scriptsize{$\pm$19.24} & 60.18\scriptsize{$\pm$13.74} & 52.44\scriptsize{$\pm$7.54} \\
\midrule
\multirow{4}{*}{GPT} & GPPT                      & 61.27\scriptsize{$\pm$2.36} & 63.53\scriptsize{$\pm$2.65} & 71.26\scriptsize{$\pm$0.62} & 53.03\scriptsize{$\pm$2.37} & 53.69\scriptsize{$\pm$\phantom{0}8.67} & 47.85\scriptsize{$\pm$\phantom{0}6.10} & 45.54\scriptsize{$\pm$8.17} \\ 
& GraphPrompt                             & 62.23\scriptsize{$\pm$1.59} & 61.75\scriptsize{$\pm$0.64} & 74.45 \scriptsize{$\pm$0.76} & 56.11 \scriptsize{$\pm$1.04} & 53.66 \scriptsize{$\pm$11.52} & 53.75 \scriptsize{$\pm$5.63} & 52.35 \scriptsize{$\pm$4.42} \\ 
& GraphPrompt+                             & 60.75\scriptsize{$\pm$1.21} & 63.21\scriptsize{$\pm$2.22} & 75.11\scriptsize{$\pm$0.24} & 54.24\scriptsize{$\pm$1.45} & 51.23\scriptsize{$\pm$9.11} & 75.11\scriptsize{$\pm$0.24} & 52.96\scriptsize{$\pm$8.26} \\ 
& GPF+                             & 70.52\scriptsize{$\pm$1.43} & 69.65\scriptsize{$\pm$1.08} & 77.45\scriptsize{$\pm$0.65} & 56.42\scriptsize{$\pm$1.31} & 70.14\scriptsize{$\pm$25.64} & 52.71\scriptsize{$\pm$22.99} & 52.61\scriptsize{$\pm$5.28} \\ 
\midrule
Ours & SGPT                          & \textbf{76.08}\scriptsize{$\pm$1.24} & \textbf{75.77}\scriptsize{$\pm$1.64} & \textbf{84.88}\scriptsize{$\pm$0.54} & \textbf{64.74}\scriptsize{$\pm$4.41} & \textbf{78.92}\scriptsize{$\pm$7.44} & \textbf{65.93}\scriptsize{$\pm$16.75} & \textbf{54.17}\scriptsize{$\pm$5.51} \\ 
\midrule
\bottomrule
\end{tabular}
\label{table:performance}
\end{table*}
\myparagraph{Baselines}
We assess the performance of SGPT against the state-of-the-art methods from four categories outlined below.
\begin{itemize}
    \item \textbf{Supervised Graph Neural Networks (GNNs)}: GCN~\cite{niepert2016learning} and GAT~\cite{velivckovic2018graph}. GCN is a pioneering and representative GNN model designed for unsigned graphs. It utilizes layer-wise message-passing algorithms to aggregate features from neighbors.
    GAT adopts the attention architecture within GNNs, allowing it to learn different aggregation weights for different neighbors.
    These unsigned GNNs are trained in a fully supervised manner. 
    \item \textbf{Supervised Signed Graph Neural Networks (SGNNs)}: SGCN~\cite{derr2018signed} and SDGNN~\cite{huang2021sdgnn}. SGCN generalizes GCN to signed graphs based on balance theory, where nodes differentially aggregate features from neighbors of distinct polarities.
    SDGNN incorporates motifs extracted from signed graphs based on balance theory and status theory, and leverages attention mechanisms for message-passing. 
    These methods are also trained in a fully supervised manner. 
    \item \textbf{GNNs with “Pre-train and Fine-tune” (FT)}: Link Prediction~\cite{hu2019linkpred} and GraphCL~\cite{you2021graph}.
    Link Prediction involves predicting the existence of masked links using the given graph structure.
    GraphCL leverages graph augmentation to generate augmented graph views and uses contrastive loss to enhance the robustness of GNNs.
    GNNs are pre-trained with Link Prediction or GraphCL and then fine-tuned during the downstream phase.
    \item \textbf{GNNs with “Pre-train and Prompt-tune” (GPT)}: GPPT~\cite{sun2022gppt}, GraphPrompt~\cite{liu2023graphprompt}, GraphPrompt+~\cite{yu2024graphprompt++}, and GPF+\cite{fang2024GPF}. GPPT unifies the downstream node classification tasks with pre-training link prediction tasks. 
    It introduces several class prototypes into graphs and predicts node labels by selecting the prototypes with the highest similarity. 
    GraphPrompt extends GPPT to subgraph-level, and applies a prompt vector to the READOUT layer of pre-trained GNNs.
    GraphPrompt+ incorporates prompt vectors within every layer of pre-trained GNN encoders and adds a weighted sum of representations learned from each layer to capitalize the hierarchical information.
    GPF+ adds attentive token vectors to the input node feature to modify the input space to align it with that of the pre-training task. 
    For these methods, GNNs are first pre-trained with Link Prediction, then the pre-trained models are frozen, and only the parameters of prompts are optimized.
\end{itemize}
\myparagraph{Implementations}
For NC tasks, we follow the few-shot setting in HGPrompt~\cite{yu2024hgprompt} and only use a small number of node instances for training. 
For LSP tasks, we follow the setting in ProG~\cite{sun2023all}. 
Here, signed links are divided into three parts to simulate the scenarios of label scarcity: 
(1) 30\% of the links are for message-passing only.
For signed graph models such as SGNNs and SGPT, these link signs are known when generating the node embeddings or applying the graph template.
For unsigned models, these links are treated as unsigned.
(2) A small number of link signs, equal to the selected shot number, are used only for model supervision.
(3) The remaining links are used for testing. 
To ensure a fair comparison regarding parameter size across different methods, we use the same GNN layer size of three and hidden dimensions of 32, 64, and 64, respectively.
We randomly generate 100 tasks for both NC and LSP tasks and report the average ROC-AUC score for evaluation.
The initial learning rate is set to 0.001, and the Adam optimizer is used.

\subsection{Performance Evaluation}
We first evaluate the performance of SGPT on few-shot NC and LSP tasks against selected baselines. 
For methods that involve pre-trained GNNs, we use GCN as the pre-training backbone. We set the shot number to 1 and 100 for NC and LSP tasks, respectively.
For SGPT, we use 2 hops to apply the graph template.
Subsequently, we vary the training shot number to analyze the performance trend. 
Afterward, we change the pre-training GNN backbones to evaluate the models' flexibility.

\myparagraph{Performance on few-shot tasks} 
We present the performance of SGPT and baselines in Table \ref{table:performance}. Based on the results, we make the following observations: 
(1) Supervised GNNs and SGNNs struggle to achieve optimal performance for both tasks, primarily due to the limited label supervision. 
In contrast, SGPT leverages transferable knowledge learned from the pre-training task and adapts the pre-trained model to different downstream tasks effectively, even when labels are scarce.
(2) For LSP tasks, LinkPred achieves better results compared to GraphCL. 
This is because LinkPred focuses on understanding the interactions between two nodes, which aligns closely with the objective of the downstream LSP. 
On the other hand, GraphCL is more effective in NC tasks as it captures the position of nodes within the entire graph. Such global features are essential for node classification. 
This shows that it is crucial to carefully select a pre-training approach that is well-aligned with the target task to ensure optimal performance.
(3) SGPT outperforms all baselines across selected datasets. 
Unlike the fine-tuning methods, SGPT effectively bridges the gap in task objectives via the task template.
Compared to unsigned prompt tuning models, i.e., GPPT, GraphPrompt, GraphPromp+ and GPF+, SGPT reduces the divergence in graph types via the graph template and captures the complex link semantic information in signed graphs, achieving the best performance.
%\vspace{-20pt}
\begin{table*}[h]
\centering
\fontsize{9pt}{10pt}\selectfont
\setlength{\tabcolsep}{1mm}
\caption{Performance with different pre-trained GNN backbones in ROC-AUC (\%). The best performance is bolded.}
\begin{tabular}{@{}cc|cccc|cccc|cccc@{}}
\toprule
\multirow{2}{*}{Task} & \multirow{2}{*}{Dataset} & \multicolumn{4}{c|}{GCN} & \multicolumn{4}{c|}{GAT} & \multicolumn{4}{c}{GIN} \\ 
\cmidrule(lr){3-6} \cmidrule(lr){7-10} \cmidrule(l){11-14}   
 & & Supervised & FT & GPF+ & SGPT & Supervised & FT & GPF+ & SGPT & Supervised & FT & GPF+ & SGPT \\ \midrule
 \midrule
\multirow{4}{*}{LSP} & Bitcoin-Alpha & 71.28  &  73.29  &  70.52  &  \textbf{76.08}  &  67.25  &  72.05  &  66.46  &  \textbf{72.95}  &  70.77  &  72.79  &  69.00  &  \textbf{75.85} \\
 & Bitcoin-Otc & 70.77  &  73.41  &  69.65  &  \textbf{75.77}  &  73.04  &  74.10  &  70.11  &  \textbf{73.34}  &  73.90  &  75.93  &  71.63  &  \textbf{76.07}\\
 & Epinions & 74.12  &  78.47  &  77.45  &  \textbf{84.88}  &  68.78  &  71.69  &  68.09  &  \textbf{78.14}  &  70.78  &  74.80  &  68.02  &  \textbf{78.36} \\
 & Slashdot & 55.48  &  57.25  &  56.42  &  \textbf{64.74}  &  56.16  &  55.73  &  56.42  &  \textbf{59.55}  &  61.15  &  63.20  &  60.78  &  \textbf{66.86} \\ \midrule
\multirow{3}{*}{NC} & WikiElec & 68.70  &  49.68  &  70.14  &  \textbf{78.92}  &  51.11  &  57.56  &  52.09  &  \textbf{65.62}  &  44.72  &  64.61  &  48.56  &  \textbf{69.49} \\
 & WikiRfA & 62.77  &  51.53  &  52.71  &  \textbf{65.93}  &  52.47  &  54.18  &  49.36  &  \textbf{55.68}  &  45.46  &  53.53  &  50.83  &  \textbf{54.35} \\
 & WikiEditor & 52.50   &  49.03   &  52.61   &  \textbf{54.17}   &  48.92   &  46.37   &  53.06   &  \textbf{57.41}   &  43.52   &  48.62   &  42.83   &  \textbf{57.41} \\
 \midrule
\bottomrule
\end{tabular}
\label{tab:backbone}
\end{table*}

\begin{figure}[t]
    \centering
    \begin{subfigure}[t]{0.49\textwidth}
        \includegraphics[width=0.9\textwidth]{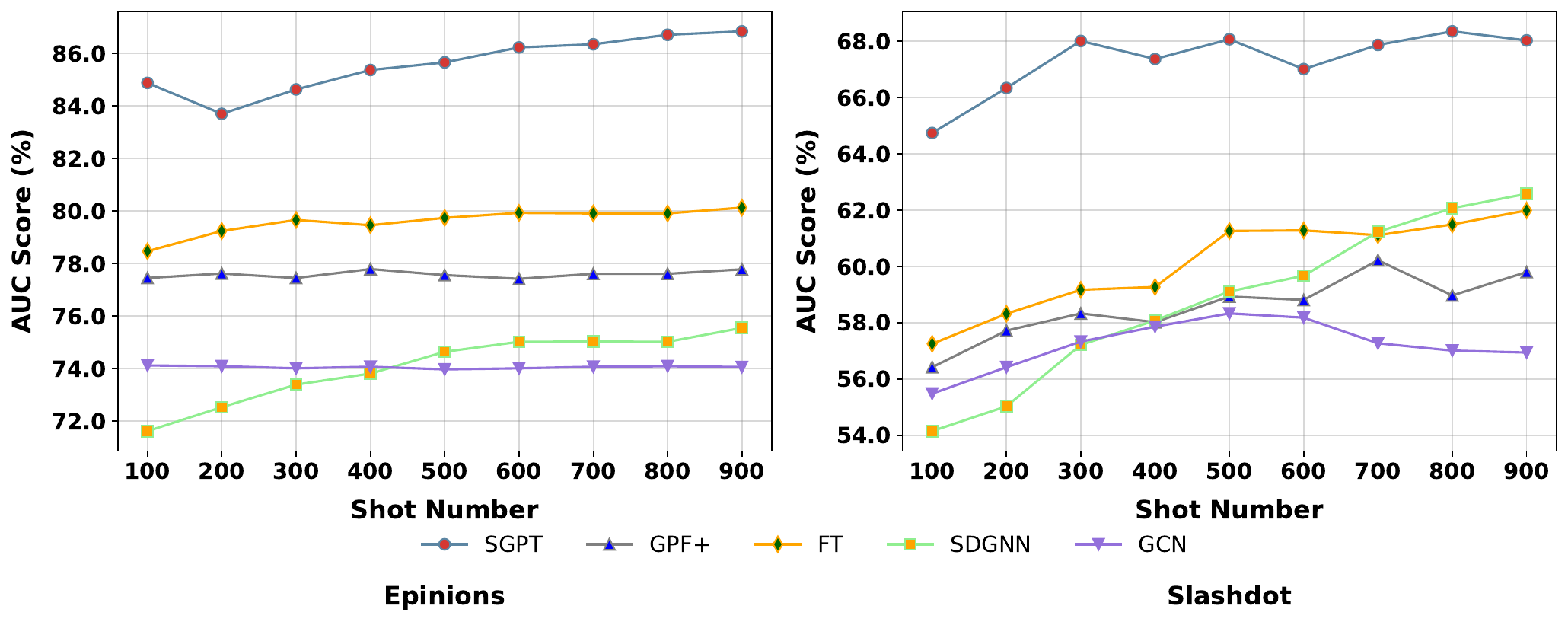}
        \caption{Link Sign Prediction.}
        \label{fig:shot_link}
    \end{subfigure}
    \hfill
    \begin{subfigure}[t]{0.49\textwidth}
        \includegraphics[width=0.9\textwidth]{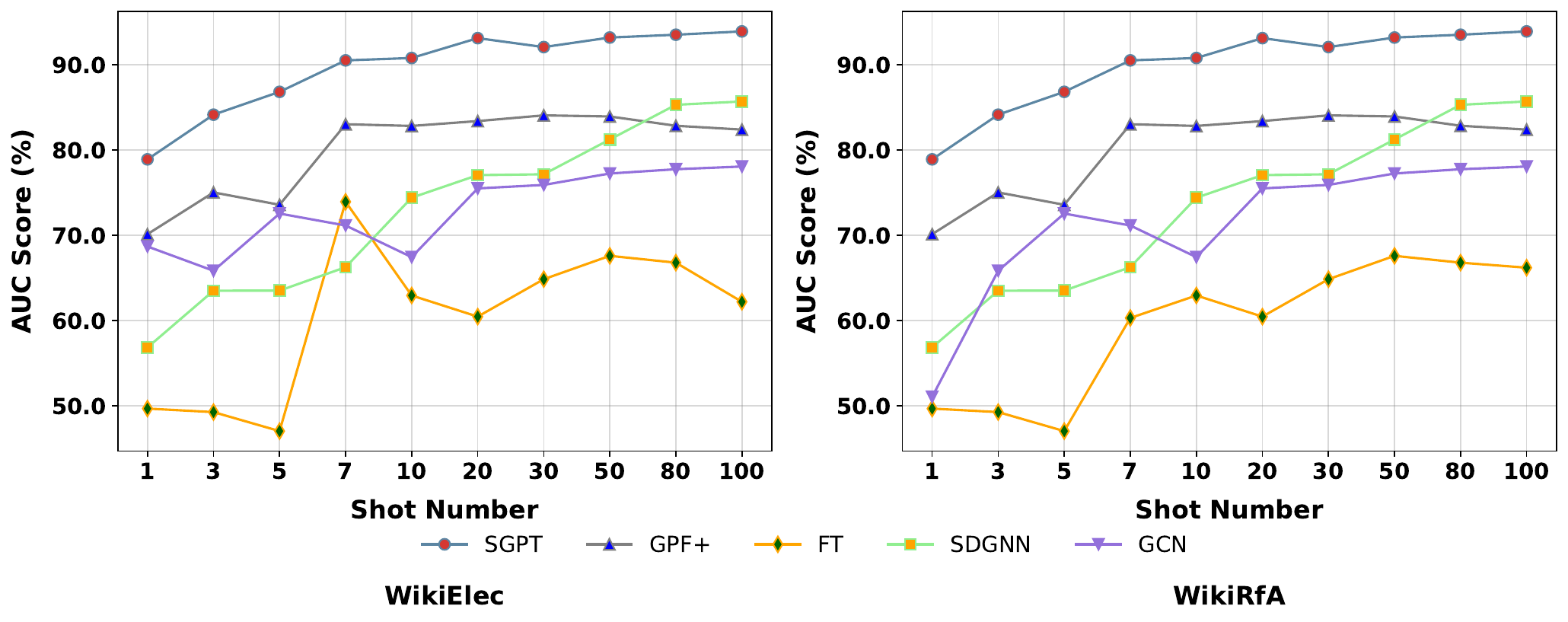}
        \caption{Node Classification.}
        \label{fig:shot_node}
    \end{subfigure}
    \caption{Impacts of shots on downstream tasks.}
\end{figure}

\myparagraph{Performance with different shots}
To further investigate the impacts of supervision shots, we evaluate the model performance with different training shots. 
We present the performance trends of SGPT against other competitive baselines in each category in Fig.~\ref{fig:shot_link} and Fig.~\ref{fig:shot_node}. 
Based on the results, we can find that (1) SGPT outperforms other baselines on both tasks when training shots are limited, and it exhibits competitive and stable performance as the shot number increases, demonstrating the effectiveness of our method.
(2) For unsigned models, e.g., GPF+, FT, GCN, no significant improvement in performance is observed as the shot number increases. This suggests that unsigned models cannot capture the complex link semantics in signed graphs, demonstrating the necessity of the proposed graph template.
(3) As the number of training shots increases, the performance of SDGNN continues to improve. 
This indicates that supervised SGNNs require adequate training labels to achieve optimal performance.

\myparagraph{Performance with different GNN backbones} 
We also assess the flexibility of SGPT with different pre-trained GNN backbones. 
We use Link Prediction to pre-train GCN, GAT, and GIN and then freeze them while prompt tuning. 
The shot number is set to 1 and 100 on NC and LSP tasks, respectively. 
We compare SGPT with supervised training (Supervised), fine-tuning (FT), and a competitive unsigned graph prompt tuning method, i.e., GPF+. 
From Table \ref{tab:backbone}, it can be observed that SGPT outperforms other methods regardless of utilized pre-trained backbones, indicating the flexibility and robustness of SGPT.
\begin{table}[t]
\centering
\small
\renewcommand{\arraystretch}{1.1}
\caption{Tunable parameters comparison.}
\begin{tabular}{l|cccccc}
\toprule
 & GCN & SDGNN & Fine-tune & GPPT & GPF+& SGPT \\
\midrule
LSP   & $\sim$15K & $\sim$20K & $\sim$15K & $\sim$13K & $\sim$0.6K & $\sim$4K \\
NC  & $\sim$8K  & $\sim$12K & $\sim$8K  & $\sim$1K  & $\sim$0.3K & $\sim$4K \\
\bottomrule
\end{tabular}
\label{tab:param-final}
\end{table}

\subsection{Ablation study} 

\begin{table*}[ht]
\centering
\fontsize{9pt}{10pt}\selectfont
\setlength{\tabcolsep}{1mm}
\caption{Performance of variants used in the ablation study in ROC-AUC score (\%). The best performance is bolded.}
\begin{tabular}{@{}c|cccc|cccc|ccc@{}}
\toprule
\multirow{2}{*}{Methods} & Task & Graph & Feature &
Semantic & \multicolumn{4}{c|}{LSP} &
\multicolumn{3}{c}{NC}\\ 
 & Template & Template & Prompt &
Prompt & Bit-Alpha & Bit-Otc & Epinions & Slashdot & WikiElec & WikiRfA & WikiEditor \\ 
\midrule
\midrule
Variant-1 & $\checkmark$&-&-&-& 63.60 & 62.07 & 72.99 & 54.36 & 59.00 & 60.15 & 50.32 \\
Variant-2 & $\checkmark$&$\checkmark$& - &-& 65.13 & 63.15 & 73.79 & 54.89 & 71.78 & 57.74 & 50.98 \\
Variant-3 &$\checkmark$&$\checkmark$&$\checkmark$&-& 71.16 & 63.55 & 73.62 & 55.18 & 71.61 & 60.83 & 50.41 \\
SGPT & $\checkmark$&$\checkmark$&$\checkmark$&$\checkmark$& \textbf{76.08} & \textbf{75.77} & \textbf{84.88} & \textbf{64.74} & \textbf{78.92} & \textbf{65.93} & \textbf{54.17}\\
\midrule
\bottomrule
\end{tabular}
\label{tab:ablation}
\end{table*}
\begin{figure*}[t]
  \centering
  \begin{subfigure}[b]{0.33\textwidth}
    \includegraphics[width=.95\linewidth]{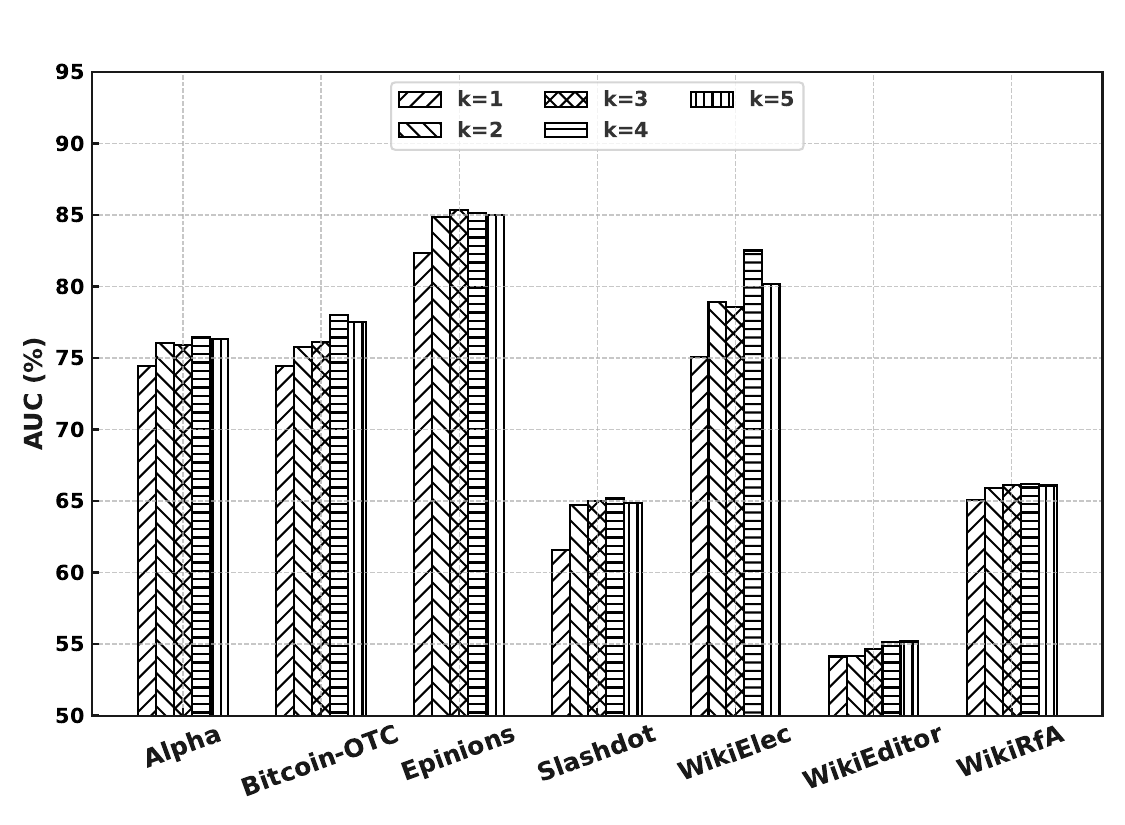}
    \caption{\footnotesize AUC-ROC with varying $k$.}
    \label{fig:auc}
  \end{subfigure}
  \hfill
  \begin{subfigure}[b]{0.33\textwidth}
    \includegraphics[width=.95\linewidth]{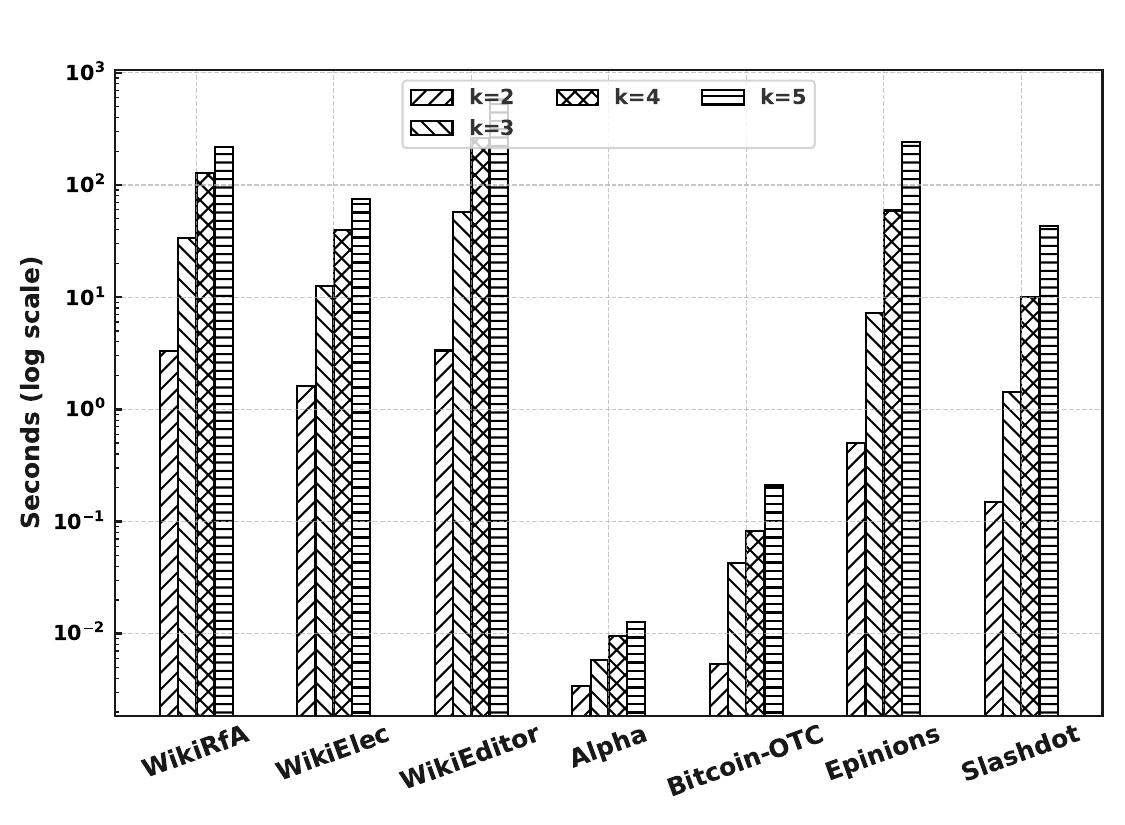}
    \caption{\footnotesize Construction time with varying $k$.}
    \label{fig:time}
  \end{subfigure}
  \hfill
  \begin{subfigure}[b]{0.33\textwidth}
    \includegraphics[width=.95\linewidth]{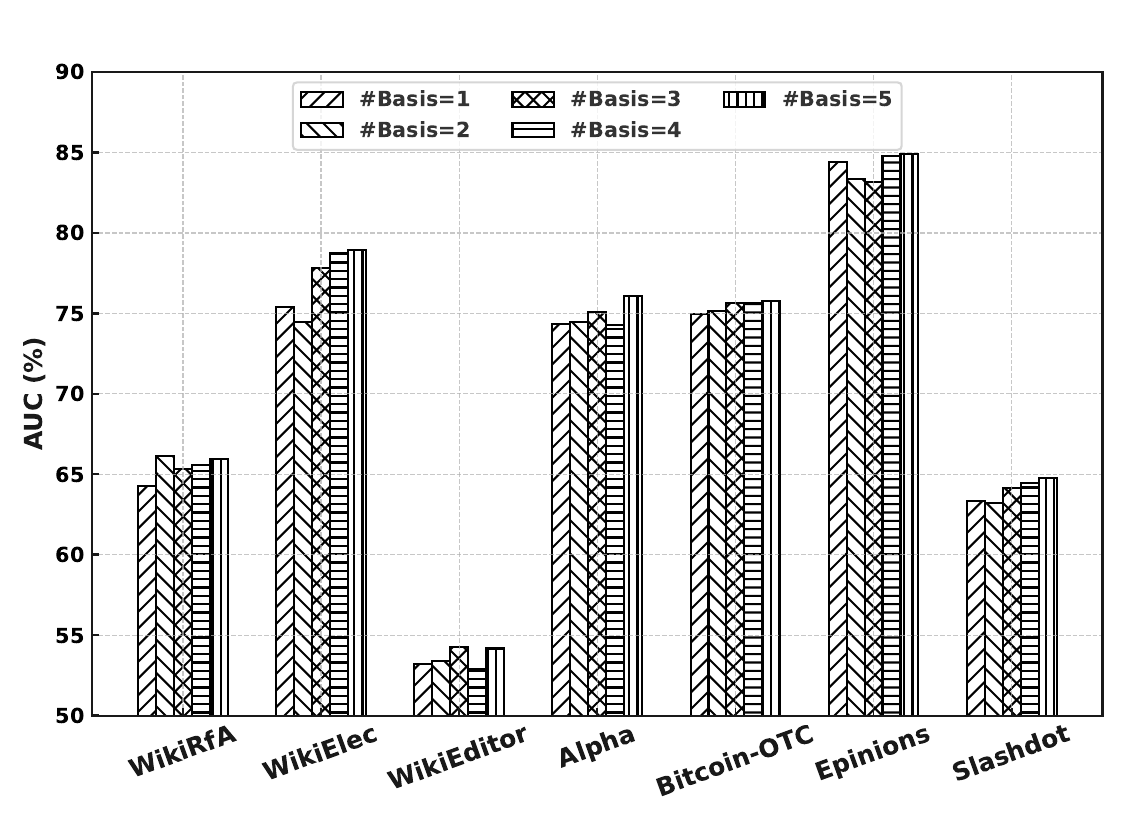}
    \caption{\footnotesize AUC-ROC with varying basis vectors.}
    \label{fig:prompt}
  \end{subfigure}
  \caption{Hyper-parameter analysis results.}
  \label{fig:combined}
\end{figure*}

To assess the contribution of each component, we compare SGPT with four variants, each lacking a component in our design.
As the task template is the basis of graph prompting, we keep it for all the variants. 
Based on the results in Table \ref{tab:ablation}, we make the following observations:
(1) Variant-3 removes the semantic prompt and adds the embedding from three channels as the final representation.
The absence of the semantic prompt significantly impacts the performance, which indicates the importance of adaptively integrating semantic information.
(2) Variant-2 further removes feature prompts. 
The model performance declines on most datasets, which demonstrates the critical role of feature alignment.
The feature prompts of SGPT are semantic-aware and effectively bridge the feature differences between pre-training and downstream tasks in each graph channel.
For datasets such as WikiElec and Epinions, the performance change is marginal, this may be attributed to inherently minimal feature drift between the pre-training and downstream stages, thereby reducing the need for explicit feature alignment.
(3) Variant-1 only retains the task template. It cannot distinguish link semantics and aggregates information from neighbors uniformly.
The poorest performance demonstrates the necessity and effectiveness of the graph template.

\subsection{Efficiency analysis}

We compare SGPT with baselines by evaluating tunable parameter sizes, as shown in Table \ref{tab:param-final}. 
When label availability is restricted, models with fewer tunable parameters are preferred for training stability and generalization ability. 
Compared with supervised methods, e.g., GCN, SDGNN, and supervised fine-tuning, SGPT has fewer tunable parameters, showing the efficiency of SGPT in few-shot scenarios.
Although SGPT has more tunable parameters than the unsigned graph prompting methods, e.g., GPPT and GPF+, it achieves the highest performance on signed tasks, demonstrating its capabilities of learning complex link semantics. 

\subsection{Complexity analysis}
% We analyze the complexity of SGPT in the following section.
% The computation consists three parts: Constructing the graph template, calculating the node embeddings with a pre-trained GNN and prompt tuning. 
% First, we construct the graph template on CPU as an offline preprocess. 
% The complexity of this part depends on the selected hop number $k$.
% Specifically, we perform sparse adjacency matrix multiplication to extract $k$-hop single-semantic neighbors in three relational channels.  
% Assuming a signed graph with $n$ nodes and an average degree of $\bar{d}$, the overall time complexity of constructing the graph template is $\mathcal{O}(n \cdot \bar{d}^k)$.  
% The second step involves computing node embeddings using a pre-trained GNN.
% The average number of each node is $\bar{d}^k)$ in graph template, then, the forward complexity of GNN is
% $\mathcal{O}(L\cdot n\cdot \bar{d}^k \cdot d_\text{out})$, where $L$ is the number of GNN layers and $d_\text{out}$ is the dimension of node embedding. 
% As we limit the hop number to be small and implement minibatch learning, this process is still manageable. 
% The third process is prompt tuning, including the computation of the feature prompt and the semantic prompt.
% The feature prompts modify each node, leading to a complexity of $\mathcal{O}(n)$.
% The semantic prompt applies an adapter, and the major computation is linear transformation, where the complexity is \(\mathcal{O}(n \cdot d_{\text{mid}} \cdot d_{\text{out})}\).

We analyse the complexity of SGPT in three stages: (i)~graph-template construction, (ii)~forward propagation of the frozen GNN, and (iii)~prompt tuning.  
Let \(n\) and \(l\) denote the numbers of nodes and signed links, respectively.
Let \(\bar d=2l/n\) be the average degree, \(L\) be the number of GNN layers, and \(d_{\text{out}}\) and \(d_{\text{mid}}\;(d_{\text{mid}}\!\ll\!d_{\text{out}})\) denote the output and bottleneck dimensions, respectively.
\textbf{(i) Graph template.}  
A single scan of the edge list partitions the signed graph into three sparse adjacency matrices for positive, negative, and unsigned links, which costs \(\mathcal{O}(l)\).
If \(k\)-hop neighbors are explicitly extracted by the graph template (by at most \(k-1\) sparse matrix multiplications), the cost is $\mathcal{O}(kl\bar d) = \mathcal{O}(kl^2/n)$.
\textbf{(ii) Frozen GNN forward process.}  
Each layer of the GNN performs sparse message passing with cost \(\mathcal{O}(l\,d_{\text{out}})\), giving a total of
\(
\mathcal{O}\!\bigl(L\,l\,d_{\text{out}}\bigr).
\)
\textbf{(iii) Prompt tuning.}  
The feature prompts calculate the attention scores over a small prompt basis set of size $r$, which costs \(r\times d_{\text{in}}\) for every node, incurring \(\mathcal{O}(n\,r\,d_{\text{in}})\).  
The semantic prompt applies a bottleneck adapter with parameters
\(d_{\text{out}}\!\times d_{\text{mid}}\) and \(d_{\text{mid}}\!\times d_{\text{out}}\) to each node embedding, costing
\(
\mathcal{O}\!\bigl(n\,d_{\text{out}}\,d_{\text{mid}}\bigr).
\)
% \textbf{Total.}  
%Summing the three parts yields
%\(
%\mathcal{O}(l)\;+\;
%\mathcal{O}\!\bigl(L\,l\,d_{\text{out}}\bigr)\;+\;
%\mathcal{O}\!\bigl(n\,r\,d_{\text{out}}\bigr)\;+\;
%\mathcal{O}\!\bigl(n\,d_{\text{out}}\,d_{\text{mid}}\bigr).
%\)
%Because \(k\), \(r\) and \(d_{\text{mid}}\) are small constants while \(d_{\text{mid}}\!\ll d_{\text{out}}\), the term \(\mathcal{O}(L\,l\,d_{\text{out}})\) dominates, and the overhead introduced by the two prompt stages is minor.

Summing the three parts yields
\(
\mathcal{O}(l)\;+\;
\mathcal{O}(kl^2/n)\;+\;
\mathcal{O}\!\bigl(L\,l\,d_{\text{out}}\bigr)\;+\;
\mathcal{O}\!\bigl(n\,r\,d_{\text{in}}\bigr)\;+\;
\mathcal{O}\!\bigl(n\,d_{\text{out}}\,d_{\text{mid}}\bigr).
\)
Since \(r\) and \(d_{\text{mid}}\) are small constants, the term \(\mathcal{O}(L\,l\,d_{\text{out}})\) dominates.
Thus, for SGPT, the dominant cost is the GNN forward propagation.
As the hop number $k$ increases, the number of links $l$ may grow exponentially, leading to an overwhelming computational overhead of SGPT. 
However, in practice, we observe that a larger $k$ does not consistently lead to better performance (see Section 5.6).  
Therefore, we typically set $k = 2$ to balance between model expressiveness and computational efficiency.
\vspace{-0.3em}
\subsection{Sensitivity analysis}
\myparagraph{Varying hop number $\boldsymbol{k}$}
We investigate the sensitivity of hop number in the proposed graph template on both NC and LSP tasks. 
The AUC-ROC scores and adjacency matrix construction time are illustrated in Fig. \ref{fig:auc} and Fig. \ref{fig:time}, respectively.
The figures show that from hop 1 to hop 2, the model performance improves across all datasets.
However, for $k>2$, the performance gains become marginal while the computational cost increases significantly due to the exponential growth of neighborhood nodes in the higher-hop adjacency matrix.
Thus, considering the trade-off between performance and computation costs, setting the hop number $k=2$ in the graph template is a reasonable value that yields strong performance with manageable complexity.

\myparagraph{Varying basis number}
We also evaluate the sensitivity with respect to the number of basis vectors in feature prompts.
Note that the number of basis vectors is the same for the three semantic channels.
The results in Fig. \ref{fig:prompt} show that increasing the number of basis vectors generally improves the performance, particularly on datasets such as WikiElec, Bitcoin-Alpha, and Epinions.
While more basis vectors offer richer representational flexibility, excessive vectors may introduce noise or unnecessary complexity for some datasets. For example, on WikiEditor, performance peaks when the number of basis vectors is 3 and slightly drops afterward.

\section{Conclusions}
In this paper, we address the challenge of transferring pre-trained GNNs to signed graph tasks and propose a few-shot prompt tuning framework, termed SGPT.
SGPT bridges the gaps caused by the divergence in graph types and task objectives between the pre-training and downstream phases.
Specifically, SGPT introduces graph and task templates to unify the structural semantics and task objectives across both stages.
In addition, SGPT involves feature prompts and semantic prompts to align the downstream semantic space with that of the pre-training task and to integrate relational semantics in a task-aware manner.
Finally, extensive experiments demonstrate the effectiveness of SGPT in low-label settings.

\newpage
\section*{GenAI Usage Disclosure}
We used OpenAI’s ChatGPT-4o to assist with code for figure visualization in this paper. Specifically, based on the experimental data, the tool was used to improve the visualization code in order to enhance the readability and overall quality of the figures.

We also used ChatGPT-4o to improve the grammar, clarity, and language quality of the manuscript, in a manner comparable to standard writing-assistance tools.
\bibliographystyle{ACM-Reference-Format}
\bibliography{sample-base}

\end{document}